# Real-Time and Continuous Hand Gesture Spotting: an Approach Based on Artificial Neural Networks

Pedro Neto, Dário Pereira, J. Norberto Pires, *Member, IEEE* and A. Paulo Moreira, *Member, IEEE*

*Abstract*— New and more natural human-robot interfaces are of crucial interest to the evolution of robotics. This paper addresses continuous and real-time hand gesture spotting, i.e., gesture segmentation plus gesture recognition. Gesture patterns are recognized by using artificial neural networks (ANNs) specifically adapted to the process of controlling an industrial robot. Since in continuous gesture recognition the communicative gestures appear intermittently with the non-communicative, we are proposing a new architecture with two ANNs in series to recognize both kinds of gesture. A data glove is used as interface technology. Experimental results demonstrated that the proposed solution presents high recognition rates (over 99% for a library of ten gestures and over 96% for a library of thirty gestures), low training and learning time and a good capacity to generalize from particular situations.

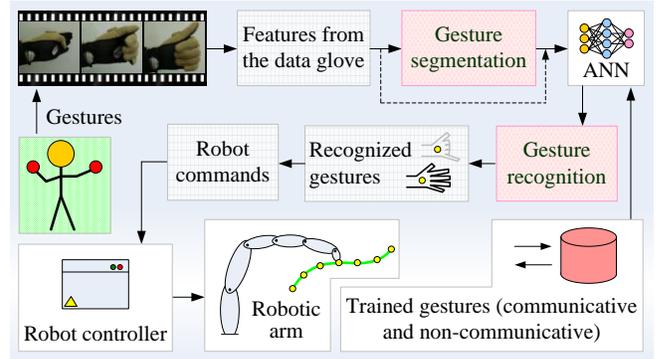

Figure 1.  The proposed system.

## I. INTRODUCTION

Reliable and natural human-robot interaction is a subject that has been extensively studied by researchers in the last few decades. Nevertheless, in most of cases, human beings continue to interact with robots recurring to the traditional process, using a teach pendant. Probably, this is because these "more natural" interaction modalities have not yet reached the desired level of maturity and reliability.

It is very common to see a human being explaining something to another human being using hand gestures. Making an analogy, and given our demand for natural human-robot interfaces, gestures can be used to interact with robots in an intuitive way. Recent research in gesture spotting (gesture segmentation plus gesture recognition) aimed at applications in many different fields, such as sign language (SL) recognition, electronic appliances control, video-game control and human-computer/robot interaction. The development of reliable and natural human-robot interaction platforms can open the door to new robot users and thus contribute to increase the number of existing robots.

### A. Interaction Technologies and Methods

Different interaction technologies have been used to capture human gestures and behaviors: vision-based systems, data gloves, magnetic and/or inertial sensors and hybrid systems combining the technologies above. Factors such as the kind of application, the cost, reliability and portability influence the choice of a technology in detriment of another.

Important work has been done in order to identify and recognize gestures using vision-based interfaces [1], for hand [2], arm [3] and full-body [4] gesture recognition. Other studies report vision-based solutions for real-time gesture spotting applied to the robotics field [5], or in the field of SL recognition [6]. An American SL word recognition system that uses as interaction devices both a data glove and a motion tracker system is presented in [7]. Another study presents a platform where static hand and arm gestures are captured by a vision system and the dynamic gestures are captured by a magnetic tracking system [8]. Inertial sensors have also been explored for different gesture-based applications [9], [10]. A major advantage of using vision-based systems in gesture recognition is the non-intrusive character of this technology. However, they have difficulty producing robust information when facing cluttered environments. Some vision-based systems are view dependent, require a uniform background and illumination, and a single person (full-body or part of the body) in the camera field of view.

Magnetic tracking systems can measure precise body motion, but at the same time, they are very sensitive to magnetic noise, expensive and need to be attached to the human body. Inertial sensors and data gloves present a number of advantages: they are relatively cheaper, allow recognizing gestures independently of the body orientation and can be used in cluttered environments. Some associated negative features are the necessity to attach them to the human body and the incapacity to extract precise displacements.

Information provided by interaction technologies have to be treated and analyzed carefully to recognize gestures. Several machine learning techniques have been used for this

*Research supported by the Portuguese Foundation for Science and Technology (FCT), PTDC/EME-CRO/114595/2009.

Pedro Neto, Dário Pereira and J. N. Pires are with the Mechanical Engineering Department, University of Coimbra, POLO II, 3030-788 Coimbra, Portugal (phone: +351239790700; fax: +351239790701; e-mail: pedro.neto@dem.uc.pt).

A. P. Moreira is with the Electrical Engineering Department, University of Porto, 4200-465 Porto, Portugal (e-mail: amoreira@fe.up.pt).

purpose, being that most of the current research in gesture recognition relies on either artificial neural networks (ANNs) or hidden Markov models (HMM) [11]. Mitra and Acharya provide a complete overview of techniques for gesture pattern recognition [1]. ANN-based problem solving techniques have been demonstrated to be a reliable tool in gesture recognition, presenting very good learning and generalization capabilities [12]. ANNs have been applied in a wide range of situations such as the recognition of continuous hand postures from gray-level video images [13], gesture recognition having acceleration data as input [14] and SL recognition [7]. The capacity of recurrent neural networks (RNN) for modeling temporal sequence learning has been demonstrated [15]. HMM are stochastic methods known for their application in temporal pattern recognition, including gesture spotting [16]. Some studies report comparisons between ANNs and HMM when they are applied to pattern recognition [17]. However, it cannot be concluded that one solution is better than the other.

*B. Proposed Approach*

This paper presents a new gesture spotting solution using a data glove as interaction technology. Gesture patterns are recognized in continuous (not separately) and in real-time recurring to ANNs specifically adapted to the process of controlling an industrial robot. Continuous gesture recognition because it is the natural way used by humans to communicate (when using gestures), in which communicative gestures (with an explicit meaning) appear intermittently with non-communicative gestures (transition gestures), with no specific order. In this way, it is proposed an architecture with two ANNs in series to recognize communicative and non-communicative gestures. Transitions between gestures are analyzed and a solution based on ANNs is proposed to deal with them. Real-time because when a user performs a gesture he/she wants to have response/reaction from the robot with a minimum delay. This takes us to the choice for static gestures rather than dynamic gestures. In fact, real-time gesture recognition imposes the use of data up to the current observation without have to wait for future data. This is not what happens when dynamic gestures are recognized as they are represented as a sequence of feature vectors.

Experimental results demonstrated that the proposed solution presents relatively good recognition rates (RRs), low training and learning time, a good capacity to generalize, it is intuitive to use, non user dependent and able to operate independently from the conditions of the surrounding environment. Fig. 1 shows a scheme of the proposed system. The data glove is a *CyberGlove II*. It has twenty-two resistive sensors for joint-angle measurements $(x_1, x_2, ..., x_{22})$ that define the hand shape in each instant of time *t*. Moreover, the glove also has a two state button.

## II. GESTURE SPOTTING

Each person can use different gestures (in this case hand gestures) to express the same desire or feeling. In this context, such gestures are associated to robot commands. To avoid ambiguities, and considering our purpose (pattern recognition), each static gesture should be different from each other. For the first experimental tests a role of ten different hand gestures (shapes) were selected, Table I. In this way, we have ten hand static gestures associated to nineteen robot commands. This is possible because we can make use of the two state button of the glove and associate the same gesture to two different robot commands just by changing the button state. After a gesture is recognized, the control command associated to that gesture is sent to the robot.

TABLE I. GESTURES AND ASSOCIATED ROBOT COMMANDS

| Gesture | Hand shape | Button ON | Button OFF |
|---|---|---|---|
| G1 | 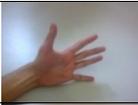 | **Stop** *The robot end-effector stops* | **Stop** *The robot end-effector stops* |
| G2 | 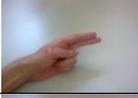 | **X+** *Motion on positive x axis* | **RX+** *Rotation about the x axis (positive direction)* |
| G3 | 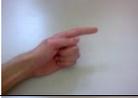 | **X-** *Motion on negative x axis* | **RX-** *Rotation about the x axis (negative direction)* |
| G4 | 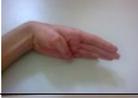 | **Y+** *Motion on positive y axis* | **RY+** *Rotation about the y axis (positive direction)* |
| G5 | 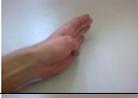 | **Y-** *Motion on negative y axis* | **RY-** *Rotation about the y axis (negative direction)* |
| G6 | 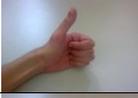 | **Z+** *Motion on positive z axis* | **RZ+** *Rotation about the z axis (positive direction)* |
| G7 | 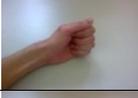 | **Z-** *Motion on negative z axis* | **RZ-** *Rotation about the z axis (negative direction)* |
| G8 | 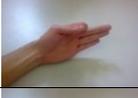 | *Save end-effector pose* | *Save end-effector pose* |
| G9 | 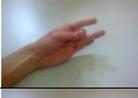 | *Return to saved pose* | *Loop* |
| G10 | 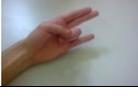 | *Vacuum ON* | *Vacuum OFF* |

*A. Gesture Segmentation*

Gesture segmentation is the task of finding the beginning and the end of a communicative gesture from continuous data. Since the duration of a gesture (static or dynamic) is variable this can be a difficult task. Several approaches have been explored to deal with the problem of gesture segmentation, some of them simply based on the definition of a threshold value, others with more complexity [18].

The proposed solution is a simple one, with the concern of achieve a spotting system with real-time characteristics.

The method consists in analysing each reading from the glove $\zeta^i = (x_1, x_2, ..., x_{22})^i$ and verify if it corresponds to a communicative gesture or not. This is possible using ANNs which are not processor intensive when classifying actual data after the training process. Nevertheless, some problems can occur, for example when during the transition from a communicative gesture to another one, a non-communicative gesture is classified as a communicative gesture. Fig. 2 shows that during the transition from *Gesture 5* to *Gesture 6* the non-communicative gestures in Fig. 2 (i) and Fig. 2 (j) can be in certain circumstances wrong classified as *Gesture 7*. This depends on the way the user performs the transition from one gesture to another. Fig. 3 shows the readings from three glove sensors ($x_1$, $x_3$ and $x_{10}$) in the scenario shown in Fig. 2. If with three sensor readings it is relatively simple to manually locate the region "*Gesture 5*", "*non-communicative gestures*" and "*Gesture 6*", for twenty-two readings the process appears more complicated. Owing to its nature, ANNs can be a good solution to deal with the scenario exposed above.

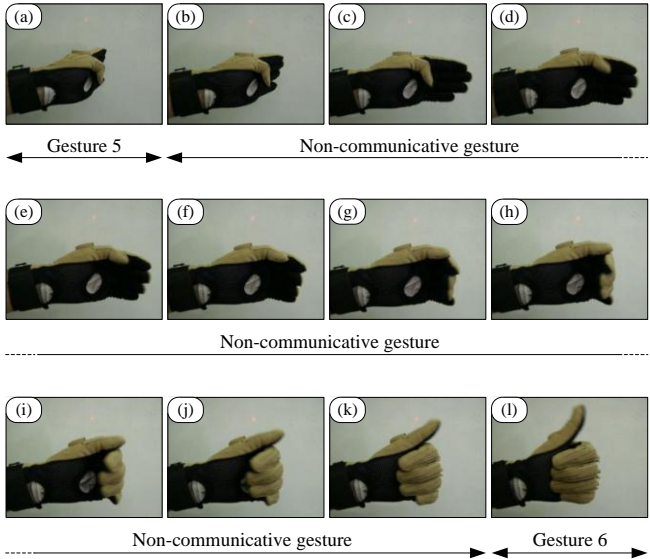

Figure 2. Transition via non-communicative gestures from *Gesture 5* to *Gesture 6*.

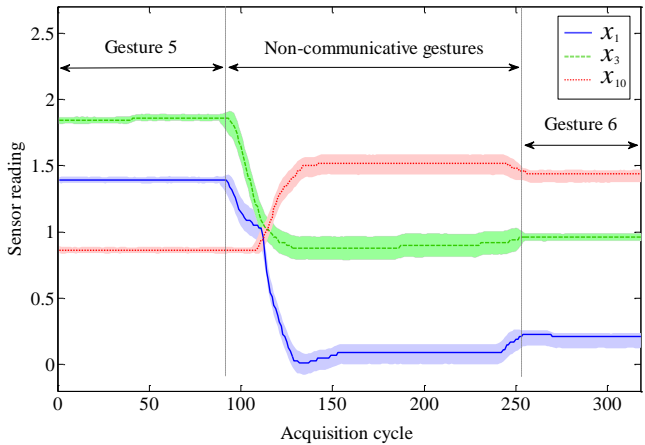

Figure 3. Data glove sensor readings in the transition from *Gesture 5* to *Gesture 6*.

## B. Gesture Recognition

Gesture recognition is the task of matching the segmented gestures against a library of predefined gestures. The main goal is to recognize gesture patterns by creating an ANN with good learning capabilities and with the ability to generalize and produce results from all kinds of input data from the glove, even if they are relatively different from the trained input patterns. The backpropagation algorithm is used as a learning/training algorithm to determine the weights of the network.

The proposed ANN architecture is a feedforward one with only one hidden layer, Fig. 4. It has forty-four neurons in the input layer, forty-four in the hidden layer and ten in the output layer. Forty-four neurons in the input layer corresponding to two consecutive readings (*t* and *t-1*) or non-consecutive (*t* and *t-n*, with $n \neq 1$) from each sensor of the glove. Since data from the glove are actualized at each 15 milliseconds this solution does not affect the real-time nature of the system if the *n* value remains low. On contrary, we have gone from a situation where gestures can be considered static to a situation in which gestures can be considered dynamic (a sequence of static gestures). In this situation the real-time character of the system can be lost. Forty-four neurons in the hidden layer because after several experiments it was concluded that this solution presents a compromise between the computational time required to train the system and the achieved RR. Finally, the ten neurons in the output layer correspond to each different gesture.

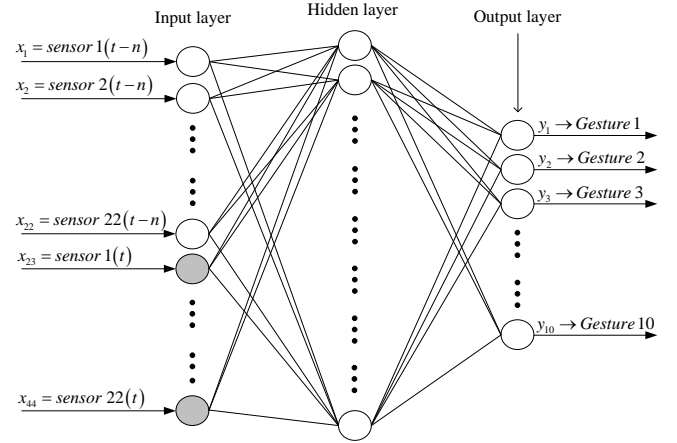

Figure 4. ANN architecture.

Considering a multi-layer ANN with *n* layers ($n \geq 3$), and being $\mathbf{y}^n$ the neurons of the last layer, $\mathbf{y}^1$ the neurons of the input layer and $\mathbf{y}^i$ the neurons of the $i^{th}$ layer. Considering also that in each layer there are *k* neurons and the desired output is *T*. Each layer can have a different activation function $\varphi^i$. The addition of the squares of the differences between the current calculated outputs and desired outputs will be the error function to minimize:

$$E = \frac{1}{2}\sum_{k=1}^{m^n}\left(T_k - y_k^n\right)^2 \qquad (1)$$

Where $m^n$ represents the number of neurons, $m$, in a layer $n$. The backpropagation algorithm is employed to find a local minimum of the error function $E$. The network is initialized with randomly chosen weights. The gradient of the error function is computed with respect to the weights **W** and bias $b$, and used to correct the initial weight values. Finally, the output $y_k^i$ of a neuron $k$ in layer $i$ is calculated by:

$$y_k^i = \varphi^i\left(v_k^i\right) = \varphi^i\left(\sum_{j=1}^{m^{i-1}} W_{k,j}^i y_j^{i-1} + b_k^i\right) \quad (2)$$

The weights can be adjusted according to the following:

$$W_{k,j}^i(n+1) = W_{k,j}^i(n) + \Delta W_{k,j}^i \quad (3)$$

$$\Delta W_{k,j}^i = -\alpha \frac{\partial E}{\partial W_{k,j}^i} = \alpha \delta_k^i y_k^{i-1} \quad (4)$$

Where $\alpha \in [0,1]$ is the learning rate. For the bias:

$$\Delta b_k^i = \alpha \delta_k^i \quad (5)$$

Being $\delta$ given by:

$$\delta_k^i = \begin{cases} \left(T_k - y_k^n\right)\dot\varphi^i\left(v_k^i\right) & \text{if } i = n \\ \left(\sum_{j=1}^{m^{i+1}} \left(\delta_j^{i+1} W_{k,j}^{i+1}\right)\right)\dot\varphi^i\left(v_k^i\right) & \text{if } 2 \leq i < n \end{cases} \quad (6)$$

The activation function is an asymmetric sigmoid:

$$\varphi(v) = \frac{1}{1+e^{-v}} \quad (7)$$

And its derivative:

$$\dot\varphi(v) = \varphi(v)\left[1-\varphi(v)\right] \quad (8)$$

The momentum is a term that can be introduced in the training of an ANN to increase the training speed and reduce instability, usually, $\beta \in [0.1,1]$. Considering that $p$ represents an iteration of the training process, the updated weight value for any connection can be calculated by the following:

$$W_p = W_{p-1} + \Delta W + \beta\left(W_{p-1} - W_{p-2}\right) \quad (9)$$

*C. Training and Recognizing Non-Communicative Gestures*

Since we are proposing an architecture with two ANNs in series to recognize communicative and non-communicative gestures (non-gestures), the system has to be trained with both kinds of gesture. This can be an important action to improve gesture segmentation by reducing false alarm situations and increasing recognition reliability. Non-gestures can be identified, trained and used to reject similar outlier patterns during gesture spotting. Non-gesture patterns are manually identified in two cases: in the transition between communicative gestures or when a false gesture is similar to a true one. This process may take a long time as it is necessary not only to identify such gestures but also to train them. Few researchers have addressed gesture recognition recurring to non-communicative gestures because it is difficult to model non-gesture patterns.

Some questions related with gesture segmentation arise: how to model non-gestures? And what should be the ANN configuration in this scenario? For the first question the answer is a simple one, non-gestures are manually identified by analyzing the transitions between communicative gestures. For the second question we propose to use an architecture with two ANNs in series, Fig. 5. Since the actual classification of a pattern is performed in few milliseconds, this solution do not affects the real-time nature of the system. These two ANNs are identical to the ANN architecture in Fig. 4. The first one classifies communicative gestures and the second one classifies non-communicative gestures. If in the first network an input pattern is not classified as a communicative gesture the process stops here, the pattern is classified as a non-communicative gesture and the system does not send any commands to the robot. On contrary, if the first network classifies an input pattern as a communicative gesture the same input pattern is used to feed the second ANN. In this case two situations can occur:

- The second ANN classifies the input pattern as a non-communicative gesture. Since it was established that this second ANN has priority over the first, we are in the presence of a non-communicative gesture.

- The second ANN does not classify the input pattern as a non-communicative gesture. In this case we are in the presence of a communicative gesture.

III. EXPERIMENTS

Experimental tests allow to evaluate the system performance in terms of RR, training time (the time the user takes to demonstrate gestures in the training phase) and computational time (the time the computer takes to train the system by adjusting the ANN weights). In a first experiment the system is tested with continuous data, in real-time and with a sequence of ten gestures (*Gesture 8*, *Gesture 2*, *Gesture 3*, *Gesture 4*, *Gesture 5*, *Gesture 6*, *Gesture 7*, *Gesture 1*, *Gesture 9* and *Gesture 10*). This sequence was chosen because it incorporates all the ten hand gestures and allows to analyze the effect of non-communicative gestures in the transition. To establish the RR for each gesture the sequence above is performed 100 times. The computer used in the experiments has a processor *Intel® Core™ 2 Duo E8400* with a memory of 1.75 GB. The ANNs are trained 10000 times, with $\alpha = 0.1$ and $\beta = 0.1$.

*A. Tests*

*Test 1:* This test was performed with the ten gestures in Table I, and recurring to an ANN having as input raw data from the glove sensors captured in two consecutive time intervals, *t-1* and *t*. Similar data (obtained similarly) are used as training patterns for the ANN, in which each different gesture is trained 20 times.

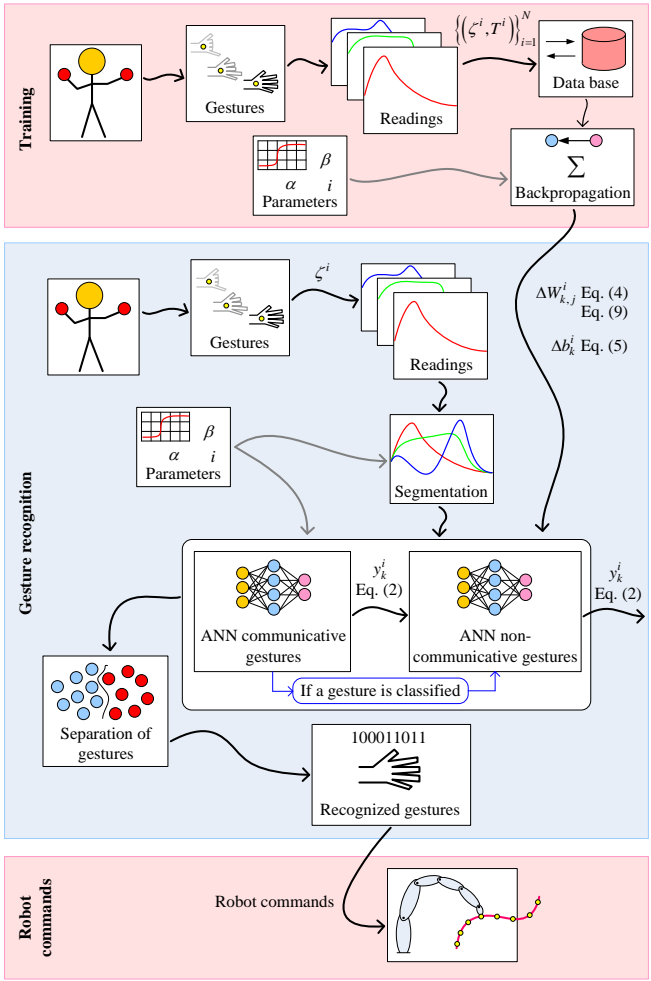

Figure 5.  System architecture.

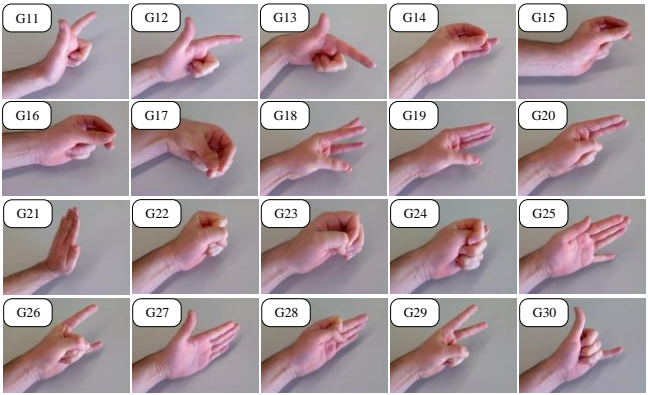

Figure 6.  Hand static gestures.

*Test 2:* Since in *Test 1* the glove sensor reading values in instants of time *t-1* and *t* are similar, we are not exploring the potentialities of the proposed ANN architecture. This test is performed having the network input patterns captured in instants of time *t-3* and *t* (45 milliseconds between these two intervals). This configuration do not affects the real-time nature of the system and allows controlling the transition from a gesture to another.

*Test 3:* Similar to *Test 2* but considering non-gestures. The network is trained with two non-communicative gestures in the transition from *Gesture 5* to *Gesture 6* and one non-communicative gesture in the transition from *Gesture 6* to *Gesture 7*.

*Test 4:* We thought it would be interesting to test the system with more gestures. This test is similar to *Test 3* (keeping the same three trained non-gestures), but, in this case for a total number of thirty gestures, ten presented in Table I more twenty gestures in Fig. 6.

*B. Results and Discussion*

For the tests presented above, the time spent in the training of the network (training time and computational time) is in Table II. The RR for each gesture is presented in Table III. In *Test 1* the system achieved an average RR of 98.4% with relatively short training time (9 minutes). Nevertheless, it was a deception to verify that *Gesture 6* is only recognized 89 times in 100. In general, in this type of situation three different errors can be pointed out:

- Substitution errors, when an input gesture is classified in a wrong category.
- Insertion errors, when the system reports a non-existent gesture.
- Deletion errors, when the system fails to detect a gesture existing in the input stream.

The low RR in *Gesture 6* occurs mainly due to substitution errors. This is because during the transition from *Gesture 5* to *Gesture 6*, sometimes, the system classifies the non-communicative gestures as to be *Gesture 7* instead *Gesture 6*, Fig. 2. In practice, when this situation occurs the user feels a small oscillation in the robot because in the transition from *Gesture 5* (Y-) to *Gesture 6* (Z+) by moments the system recognizes *Gesture 7* (Z-) and the robot reacts to that event. This issue can be solved by imposing a minimum time period that each communicative gesture should be active.

In relation to *Test 2*, the system achieved an average RR of 99.3% (96% for Gesture 6) with a total training time of 11 minutes. The training time is increased when compared to Test 1. However, the RR appears to be excellent.

In *Test 3* the system achieved an average RR of 99.8% with a total training time of 15 minutes. Thus, this ANN-based solution using models of non-communicative gestures improves gesture spotting (by reducing false alarm gestures) and increases the recognition reliability.

Finally, in *Test 4*, we have a global RR of 96.3% for a set of thirty static gestures recognized in continuous mode and in real-time. This is a very good result when compared with similar studies in the field [7], [18]. If we are dealing with a relatively high number of gestures (and non-gestures) the selection and training of non-gestures can be a very difficult task. Only a correct identification of non-gestures (ensuring that non-communicative gestures are not similar to communicative gestures) can improve the overall RR of the system. On contrary, the system can behave worse than when only communicative gestures are trained and recognized. Fig. 7 shows the robot end-effector being controlled by means of gestures.

TABLE II. TIME SPENT IN THE TRAINING PROCESS

| Process ↓ | Time [minutes] | | | |
|---|---|---|---|---|
| | *Test 1* | *Test 2* | *Test 3* | *Test 4* |
| Training time | 5 | 5 | 6 | 20 |
| Computational time | 4 | 6 | 9 | 140 |
| Total | 9 | 11 | 15 | 160 |

TABLE III. RR FOR EACH DIFFERENT GESTURE

| Gesture ↓ | Recognition rate [%] | | | |
|---|---|---|---|---|
| | *Test 1* | *Test 2* | *Test 3* | *Test 4* |
| G8 | 100.0 | 100.0 | 100.0 | For a library of thirty different hand gestures, Table I and Fig. 6 |
| G2 | 100.0 | 100.0 | 100.0 | |
| G3 | 100.0 | 100.0 | 100.0 | |
| G4 | 98.0 | 100.0 | 100.0 | |
| G5 | 98.0 | 100.0 | 100.0 | |
| G6 | 89.0 | 96.0 | 98.0 | |
| G7 | 100.0 | 97.0 | 100.0 | |
| G1 | 100.0 | 100.0 | 100.0 | |
| G9 | 99.0 | 100.0 | 100.0 | |
| G10 | 100.0 | 100.0 | 100.0 | |
| Mean | 98.4 | 99.3 | 99.8 | 96.3 |

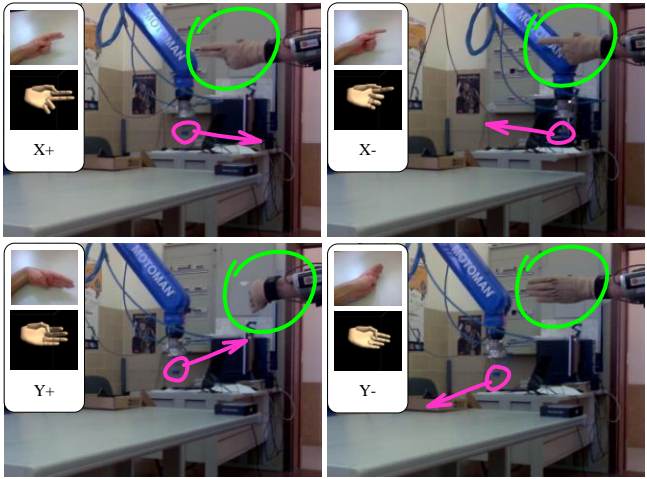

Figure 7. Examples of robot motion actions controlled by means of hand gestures.

## IV. CONCLUSIONS AND FUTURE WORK

A method for real-time and continuous hand gesture spotting has been presented. The proposed solution allows users to teach robots in an intuitive way, using gestures. Gesture patterns are classified using ANNs, which can be trained with communicative and non-communicative gestures. Experimental results report very good RRs (99.8% for a library of ten gestures and 96.3% for a library of thirty gestures), low training and learning time, a good capacity to generalize, and ability to operate independently from the conditions of the surrounding environment.

Future work will seek to improve the achieved RR. This can be done by adding more components to the actual interaction technologies, the data glove can be combined with inertial sensors or a magnetic-based tracking system. New methods dedicated to the automatic generation of non-gestures will be studied, especially an approach using random gestures as non-communicative gestures.

REFERENCES

[1] S. Mitra and T. Acharya, "Gesture recognition: a survey," *IEEE Trans. Systems, Man Cybernetics*, vol. 37, no. 3, pp. 311–324, 2007.

[2] H. Francke, J. R. del Solar and R. Verschae, "Real-time hand gesture detection recognition using boosted classifiers and active learning," in *Proc. Pacific Rim Adv. Image and Video Tech.*, 2007, pp. 533–547.

[3] T. Kirishima, K. Sato and K. Chihara, "Real-time gesture recognition by learning and selective control of visual interest points," *IEEE Trans. on Pattern Analysis and Machine Intelligence*, vol. 27, no. 3, pp. 351–364, 2005.

[4] D. Weinland, R. Ronfard and E. Boyer, "Free viewpoint action recognition using motion history volumes," *Computer Vision and Image Understanding*, vol. 104, no. 2/3, pp. 249–257, 2006.

[5] I. Mihara, Y. Yamauchi and M. Doi, "A real-time vision-based interface using motion processor and applications to robotics," *Systems and Computers in Japan*, vol. 34, no. 3, pp. 10–19, 2003.

[6] G. Lalit and M. Suei, "Gesture-based interaction and communication: automated classification of hand gesture contours," *IEEE Trans. Systems, Man and Cybernetics*, vol. 31, no. 1, pp. 114–120, 2001.

[7] C. Oz and M. C. Leu, "Linguistic properties based on American sign language isolated word recognition with artificial neural networks using a sensory glove and motion tracker," *Neurocomputing*, vol. 70, no. 16/18, pp. 2891–2901, 2007

[8] M. Strobel, J. Illmann, B. Kluge and F. Marrone, "Using spatial context knowledge in gesture recognition for commanding a domestic service robot," in *Proc. 11th IEEE International Symposium on Robot and Human Interactive Communication*, 2002, pp. 468–473.

[9] L. Kratz, M. Smith and F. J. Lee, "3D gesture recognition for game play input," in *Proc. Conference on Future Play*, 2007, pp. 209–212.

[10] P. Neto, J. N. Pires and A. P. Moreira, "Accelerometer-based control of an industrial robotic arm," in *Proc. 18th IEEE International Symposium on Robot and Human Interactive Communication*, Toyama, 2009, pp. 1192–1197.

[11] F. Parvini, D. McLeod, C. Shahabi, B. Navai, B. Zali and S. Ghandeharizadeh, "An approach to glove-based gesture recognition," in *Human-Computer Interaction. Novel Interaction Methods and Techniques*, J. Jacko, Ed. Berlin: Springer, 2009, pp. 236–245.

[12] K. Madani, "Industrial and real world applications of artificial neural networks illusion or reality?," in *Informatics in Control, Automation and Robotics I*, J. Braz et al., Ed. 2006, pp. 11–26.

[13] C. Nolker and H. Ritter, "Visual recognition of continuous hand postures," *IEEE Trans. Neural Networks*, vol. 13, no. 4, pp. 983–994, 2002.

[14] J. Yang, W. Bang, E. Choi, S. Cho, J. Oh, J. Cho, S. Kim, E. Ki and D. Kim, "A 3D hand-drawn gesture input device using fuzzy ARTMAP-based recognizer," *J. of Systemics, Cybernetics and Informatics*, vol. 4, no. 3, pp. 1–7, 2006.

[15] Y. Yamashita and J. Tani, "Emergence of functional hierarchy in a multiple timescale neural network model: a humanoid robot experiment," *PLoS Comput. Biol.*, vol. 4, no. 11, pp. 1–17, 2008.

[16] B. Peng and G. Qian, "Online gesture spotting from visual hull data," *IEEE Trans. on Pattern Analysis and Machine Intelligence*, vol. 33, no. 6, pp. 1175–1188, 2011.

[17] H. H. Avilés, W. Aguilar and L. A. Pineda, "On the selection of a classification technique for the representation and recognition of dynamic gestures," in *IBERAMIA 2008*, H. Geffner et al., Ed. Berlin-Heidelberg: Springer, 2008, pp. 412–421.

[18] H. D. Yang, S. Sclaroff and S. W. Lee, "Sign language spotting with a threshold model based on conditional random fields," *IEEE Trans. on Pattern Analysis and Machine Intelligence*, vol. 31, no. 7, pp. 1264–1277, 2009.